%% file: main.tex
\documentclass[conference]{IEEEtran}
\usepackage{times}
\usepackage{color}
\usepackage{cuted}
\usepackage{amsfonts,amsmath,amssymb,amsthm}
\usepackage{caption}
\usepackage{booktabs}
\usepackage{graphicx}
\usepackage{changes}
\usepackage{algorithm}
\usepackage{algorithmic}

\usepackage[numbers]{natbib}
\usepackage{multicol}
\usepackage[bookmarks=true]{hyperref}
\usepackage[utf8]{inputenc}
\usepackage{graphicx}

\begin{document}

\title{Broadcasting Support Relations Recursively from Local Dynamics for Object Retrieval in Clutters}

% You will get a Paper-ID when submitting a pdf file to the conference system
\author{
Yitong Li$^{* 1,2}$ \quad
Ruihai Wu$^{* 1,4}$ \quad \\ 
Haoran Lu$^{1,4}$ \quad 
Chuanruo Ning$^{1,4}$ \quad 
Yan Shen$^{1,4}$ \quad 
Guanqi Zhan $^{3}$ \quad
Hao Dong $^{1,4}$ \quad  \\
$^1$CFCS, School of CS, PKU \quad 
$^2$Weiyang College, THU \quad 
$^3$University of Oxford
\\$^4$National Key Laboratory for Multimedia Information Processing, School of CS, PKU
}

%\author{\authorblockN{Michael Shell}
%\authorblockA{School of Electrical and\\Computer Engineering\\
%Georgia Institute of Technology\\
%Atlanta, Georgia 30332--0250\\
%Email: mshell@ece.gatech.edu}
%\and
%\authorblockA{Twentieth Century Fox\\
%\and
%\authorblockN{James Kirk\\ and Montgomery Scott}
%\authorblockA{Starfleet Academy\\
%San Francisco, California 96678-2391\\

% avoiding spaces at the end of the author lines is not a problem with
% conference papers because we don't use \thanks or \IEEEmembership

%Homer Simpson\authorrefmark{2},
%Montgomery Scott\authorrefmark{3} and
%Eldon Tyrell\authorrefmark{4}}
%Telephone: (800) 555--1212, Fax: (888) 555--1212}
%\authorblockA{\authorrefmark{4}Tyrell Inc., 123 Replicant Street, Los Angeles, California 90210--4321}}

\maketitle
\renewcommand{\thefootnote}
{\fnsymbol{footnote}}
\footnotetext[1]{Equal contribution.}

% \begin{center}
%     \centering
%     \captionsetup{type=figure}
%     \includegraphics[width=1 \linewidth]{figs/fig1.png}
%     \caption{ roughly show process}
%     \label{fig:galaxy}
% \end{center}

\begin{strip}
\vspace{-10mm}
    \centering
    \includegraphics[width=\linewidth]{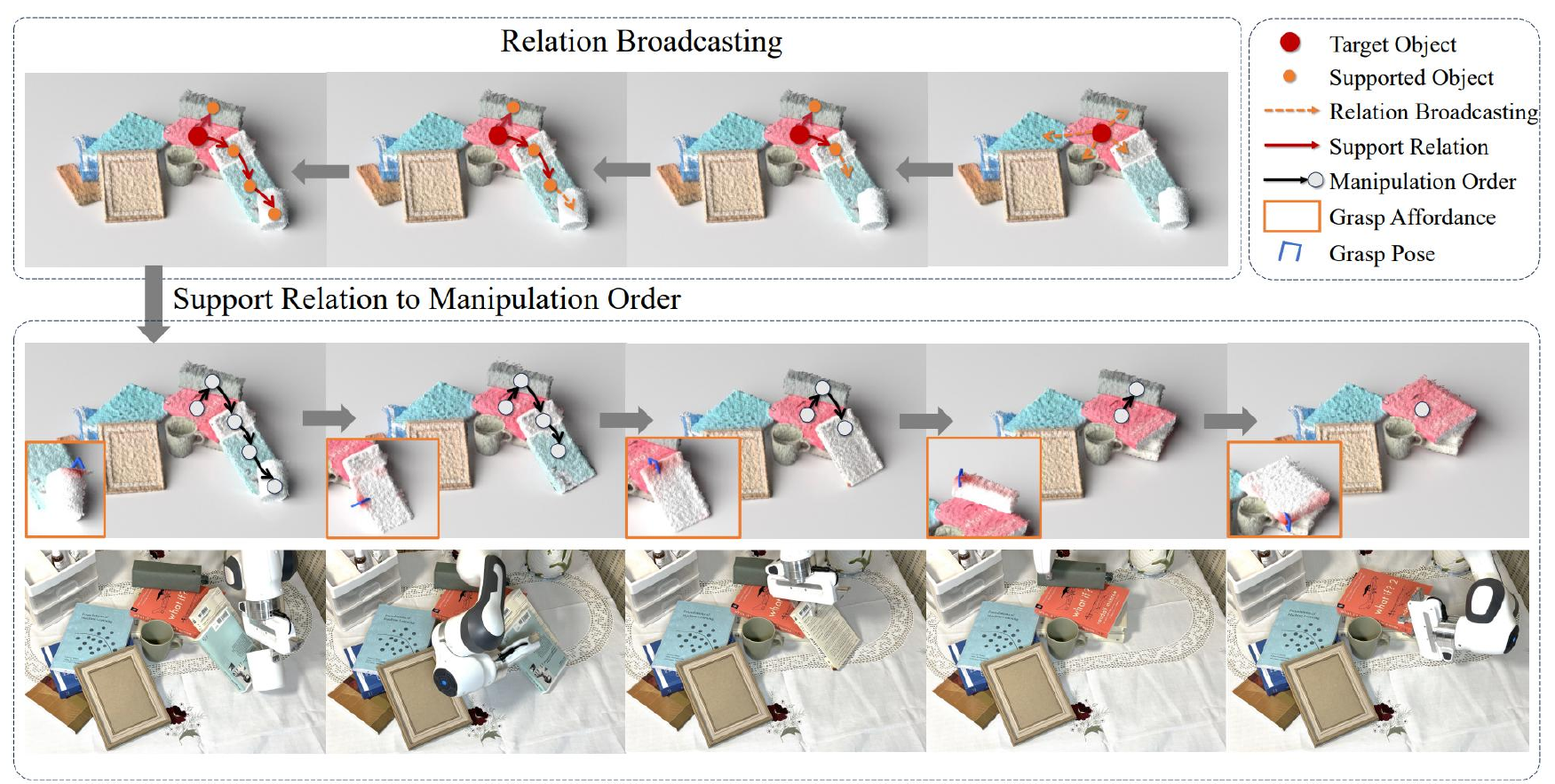}
% \vspace{-10mm}
    \captionof{figure}{\small
    \textbf{Our Proposed Framework} broadcasts the support relations recursively from the target object using local dynamics between adjacent objects, and uses the support relation graph to efficiently guide the step-by-step target object retrieval.}
    \label{fig:teaser}
\end{strip}

\input{tex/0_abs}

\IEEEpeerreviewmaketitle

\input{tex/1_introduction}

\input{tex/2_related}

\input{tex/3_problem}
\input{tex/4_method}
\input{tex/5_experiment}
\input{tex/6_conclusion}

\bibliographystyle{plainnat}
\bibliography{references}
\end{document}

%% file: tex/0_abs.tex
\begin{abstract}
% Cluttered objects
% such as stationeries and books on the table, and bowls and plates in the kitchen sink, exist everywhere in our daily life.

In our daily life, cluttered objects are everywhere, from scattered stationery and books cluttering the table to bowls and plates filling the kitchen sink.
Retrieving a target object from clutters is an essential while challenging skill for robots,
for the difficulty of safely manipulating an object without disturbing others, which requires the robot to plan a manipulation sequence and first move away a few other objects supported by the target object step by step.
However, due to the diversity of object configurations (\emph{e.g.}, categories, geometries, locations and poses) and their combinations in clutters, 
it is difficult for a robot to accurately infer the support relations between objects faraway with various objects in between.
In this paper,
we study retrieving objects in complicated clutters via a novel method of recursively broadcasting the accurate local dynamics to build a support relation graph of the whole scene, 
which largely reduces the complexity of the support relation inference and improves the accuracy. 
Experiments in both simulation and the real world demonstrate the efficiency and effectiveness of our method.
% which significantly improves the manipulation successful rate while largely reduces the number of manipulated objects and times of reasoning.
% With such relations, we can easily train the affordance map and action proposals for safe manipulation. 
% As there may exist diverse objects with diverse configurations in a clutter,
% we inference the support relations in a progressive manner, from the target object to level-by-level related objects, largely reducing the complexity of the support relation inference.
% Conditioned on 
% Manipulating objects in clutter is an essential but challenging task for robotic manipulation. It is essential because there are many situations where objects are placed in clutter. 
% However, manipulating such objects requires reasoning about the support relations between different objects, which is challenging for robotic systems nowadays. In our paper, we formulate the new problem of grasping objects in clutter without breaking other objects, and set up a new evaluation benchmark for the task covering a large diversity of objects in realistic environment. A novel system based on interaction is proposed to infer
% the support relations, after which the grasping affordance for the target object is predicted conditioned on the predicted support relations. 

\end{abstract}

%% file: tex/1_introduction.tex
\section{Introduction}
Cluttered objects~\cite{wang2021graspness, goyal2022ifor, sundermeyer2021contact}, such as piled books on desks and cluttered objects in kitchen, 
are everywhere in our daily life.
To retrieve a target object in the complicated clutter~\cite{xu2023joint,huang2021visual,kurenkov2020visuomotor,zeng2018learning}, \emph{e.g.},  retrieving a book from clutters of books and stationeries, is an essential capability for future robots to assist human in various scenarios.

Compared to object-centric manipulation tasks (such as grasping a bottle, opening a drawer or closing a door),
cluttered objects manipulation like retrieving is much more challenging for many reasons.
One of the most important reasons is that, it requires safe manipulation,
which means when manipulating the target object, other objects should not be collided.
For example,
to retrieve a plate in the sink filled with plates, bowls and glasses,
the robot should avoid making other objects broken.

To achieve this goal,
the manipulation will be long-horizon. More specifically,
the robot needs to have the planning ability to first move a few other objects away step by step,
and then manipulate the target object.
For the object retrieval task in the clutters,
the relations that the robot should be aware of, is the supporting relations between objects.
That's to say,
to retrieve a target objects,
other objects that are directly or indirectly supported by it should be first moved away safely.

However,
due to the complexity of the clutters in terms of various combinations of diverse categories and shapes of objects in different positions and poses, with further existence of occlusion problem~\cite{zhan2022tri, zhan2023amodal, price2019inferring, bejjani2021occlusion},
it is very difficult to accurately infer support relations in the complicated scene and draw an accurate and efficient manipulation plan based on that.
For example,
when the two objects are distant with many objects in between,
% when pushing or grasping one object in between,
it is very difficult for a network to predict the dynamics of the other object,
as the dynamics is transferred complicatedly by the chained objects in between.

To tackle this problem,
we leverage the property of dynamics models that,
local dynamics is much more accurate and easier to predict than the dynamics between two distant objects,
and thus propose to infer to broadcast the support relations recursively from the target objects to more and more faraway objects.
More specifically,
we first infer the support relations between target and its adjacent object using the accurate local dynamics predictor.
When some of these objects are inferred to be supported by the target object,
we execute the local dynamics predictor on these objects and find new objects supported by them, which are also indirectly supported by the target object.
We apply this process recursively and gradually build the support relation graph containing all the objects that should be retrieved before retrieving the target (Figure~\ref{fig:teaser}, first row).
From this directed acylic graph (DAG),
we can easily retrieve objects from those non-outdgree objects in the graph (Figure~\ref{fig:teaser}, second the third row).

To evaluate our framework in cluttered object retrieval,
especially in complicated clutters,
while previous works on supporting relation inference for robotic manipulation use environments with relatively simple object geometries~\cite{mojtahedzadeh2015support,kartmann2018extraction,paus2021probabilistic,motoda2023multi,huang2023planning} or in the lack of clutters complexities ~\cite{panda2013learning,panda2013learning2,panda2016single,xiong2021geometric},
we propose a new evaluation environment with combinations of thousands of different objects into realistic scenarios.
In this environment, extensive experiments demonstrate that
our proposed framework outperforms previous works that directly infer object relations by a large margin.

In short, in this paper,
we make the following contributions:
\begin{itemize}
    % \item \guanqi{Can we say new evaluation benchmark to better formulate the problem?}
    \item we propose to leverage the accurate local dynamics and broadcast it recursively to study object support relations for the retrieval task in clutters;
    \item we propose a novel system with novel designs to efficiently build the support relation graphs that can guide downstream object retrieval task;
    \item extensive experiments on diverse realistic scenarios and comprehensive metrics demonstrate the superiority of our proposed framework;
\end{itemize}

% To summarise, in this paper, we make the following four contributions: (i) A new formulation of manipulating objects in clutter considering support relations is proposed, with a new evaluation benchmark for the task covering diverse objects and complicated geometry. (ii) We develop a novel system to infer support relations via data collected by physical interaction in the simulator, and estimate the manipulation affordance conditioned on the support relation predictions. (iii) Extensive experiments in both synthetic and real environments demonstrate the effectiveness and efficiency of our system, with the grasping successful rate largely improved, both the number of manipulated objects and times of reasoning significantly reduced.

%% file: tex/2_related.tex
\section{Related Work}

\subsection{Support Relations Inference}

% \guanqi{Will refine tomorrow morning}
Inferring support relations~\cite{silberman2012indoor, guo2013support,huang2015support,nie2018semantic, paus2021probabilistic,motoda2023multi} is important in object relation inference~\cite{lu2016visual, zhang2017visual, wu2020localize, liu2020beyond} in computer vision and robotics community. 
 A dataset~\cite{silberman2012indoor} manually annotates support relation of different objects or regions for RGBD images.
Following this, a series of works aim to infer support relations given different forms of input, including RGB~\cite{zhan2023does,zhuo2017indoor,nie2018semantic,nie2020shallow2deep,nie2021content} or RGBD~\cite{guo2013support,xue2015towards,yang2017support,zhang2020support} images, and 3D models~\cite{huang2015support}.
In contrast, robotic community formulates support relation inference problem based on both object relations and manipulation. 
However, object geometries~\cite{mojtahedzadeh2015support,kartmann2018extraction,paus2021probabilistic,motoda2023multi,huang2023planning} and clutter complexities ~\cite{panda2013learning,panda2013learning2,panda2016single,xiong2021geometric} are relatively simple.
Our study infers the support relations for manipulation via the broadcasting of dynamics models, and proposes a new evaluation benchmark consisting of more diverse objects with more complicated geometry.

\subsection{Cluttered Objects Manipulation}
In the realm of robotic manipulation, addressing the challenge of interacting with objects in cluttered scenes has garnered significant attention due to its practical implications for real-world applications ranging from object grasping~\cite{sundermeyer2021contact,wang2021graspness,zeng2022robotic,zhang2023gamma,chen2023efficient,lu2023vl,dai2023graspnerf,qin2020s4g,breyer2021volumetric,fang2020graspnet,ZhangLu-RSS-21}, retrieval~\cite{xu2023joint,huang2021visual,kurenkov2020visuomotor,zeng2018learning,danielczuk2019mechanical,yang2020deep,Wada2022SafePickingLS}, to rearrangement~\cite{goyal2022ifor,cheong2020relocate,lee2019efficient,tang2023selective}. 
% While certain previous work~\cite{} employed oriented rectangles in images to indicate manipulation poses, these plannar-based predictions often result in limited Degrees of Freedom (DoF). 
% Conversely, other works tackled the manipulation of cluttered objects by learning manipulation poses with full DoF, allowing for broader applicability across diverse scenes and tasks.
Some works leverage visual grounding~\cite{xu2023joint,lu2023vl,ZhangLu-RSS-21} or object detection~\cite{li2023mobile,schwarz2018rgb} technique to comprehend cluttered scenes,
while others advocate an end-to-end framework for direct manipulation pose prediction~\cite{breyer2021volumetric,fang2020graspnet,ni2020pointnet++,qin2020s4g,sundermeyer2021contact}. Among them, a subset of works~\cite{wu2023learning,wang2021graspness,chen2023efficient, wang2021adaafford, li2024unidoormanip} focuses on learning manipulation affordance as guidance for subsequent manipulation pose generation, significantly enhancing both efficiency and accuracy.
However, prior works lack explicitly consideration for the relationships within cluttered scenes, and hardly impose constraints on the movement of other objects.
Our study proposes the use of graphs to represent cluttered scenes and support relations between objects, which aids in safe object manipulation without disturbing others.
% Recognizing the importance of maintaining relative object positions, we propose the use of graphs as a suitable method for representing and understanding cluttered scenes without disturbing the placement of other objects.

% ~\cite{ZhangLu-RSS-21}

\subsection{Dynamics Models for Robotic Manipulation}
Building dynamics models has been a promising approach in robot systems, which plans the manipulation through predicting the future state of objects under different actions. Finding the suitable representations for different objects is the core task in these model-based methods ~\cite{xue2023neural, wang2023dynamic, chen2023predicting, sanchez2020learning}. Previous dynamics models made advancements by discovering appropriate abstraction for different objects, such as particle representations for deformable objects ~\cite{shi2022robocraft, shi2023robocook, li2018learning, pfaff2020learning, ma2021learning, luunigarment, CVPR} 
 and adjacent object interactions~\cite{chen2023predicting}, video predictions for rigid objects~\cite{ko2023learning, finn2017deep, xie2019improvisation, yi2019clevrer, bear2021physion}, pixel representations for granular objects ~\cite{wang2023dynamic, suh2021surprising}.
However, the suitable form of dynamics models for cluttered objects are still under-explored. Our method devises the particle representation for modeling the movement of each object and uses graph representation to model the relation among cluttered objects, which effectively establishes the dynamics model for manipulation.

%% file: tex/3_problem.tex
\section{Problem Formulation}

% Given a clutter of $n$ objects $O_1,O_2,O_3,\dots,O_n$, we want the robot to extract an object $O_i$ in the clutter safely. We formulate an object extraction task in clutter as follow: for a target object $O_m$, propose an manipulation sequences $\{A_{k_1},A_{k_2}\dots A_m\}$, where $A_{k_j}$ means $(O_{k_j},p_{k_j},P_{k_j},D_{k_j})$, a grasp action consist of an target object $O_{k_j}$, target grasp point $p_{k_j}$, grasp pose $P_{k_j}$ and extraction direction $D_{k_j}$, and $k_1,k_2\dots l$ means the object extraction order.
% \ruihai{how to define safe?}
% \ruihai{is extract the accurate word? could we use other words like retrieve? do we have reference papers discussing the word?}
Given a partially scanned 3D point cloud $S \in \mathbb{R}^{N \times 3}$ of a clutter containing $n$ objects $O_1,O_2,\dots,O_n$ (where $O_j \in \mathbb{R}^{N_j \times 3}$ for each object $O_j$), with a target object $O_t$ ($t \in \{1,2,\dots,n\}$),
the goal for the robot is to sequentially remove occluded objects that are directly or indirectly supported by the target object $O_t$, and finally retrieve $O_t$ from the clutter. Each manipulation action should be safe, meaning that the manipulation of one object should not result in displacements of other objects in their positions and poses.
% safely retrieve $O_t$ from this clutter. 
% , and each retrieval action should be safe and not disturb other objects. 

% Specifically, 
% in the time step $i$, the robot takes in the current scene $S_i$ and executes the manipulation actions $a_i$. To retrieve the target object $O_t$, the robot executes a sequence of $l$ manipulation actions $(a_1,a_2\dots a_m)$ that first sequentially remove $l-1$ objects directly or indirectly supported by the target object $O_t$, and then take out $O_t$ in the step $l$.
% Each action $a_i$ is represented as $(o_i, p_i,P_i,D_i)$, where $o_i \in \{1, 2, \dots, n\}$ with $O_{o_m} = O_{t}$ represents the index of the object for manipulation, $p_i \in \mathbb{R}^{3}$ is the grasp point on the object $O_{o_i}$ to be seized, $P_i \in SO(3)$ is corresponding grasp pose on $p_i$, and $D_i \in SO(3)$ is the manipulation orientation for retrieving $O_{o_{i}}$ after grasping on $p_i$.
% The manipulation sequence should be safe,
% meaning when manipulating each $O_{o_i}$ with $a_i$,
% other objects should not have displacements in their positions and poses.

Specifically,
at time step $i$, the robot takes in the current scene $S_i$ and executes a manipulation action $a_i$. Each action $a_i$ safely takes out one object $O_{k_i}$, where $k_i \in \{1, 2, \dots, n\}$ denotes the index of the manipulated object. 
The overall manipulation sequence involves the robot executing actions $(a_1,a_2, \dots, a_{l-1})$ in the initial $l-1$ steps to sequentially remove $l-1$ occluded objects supported by the target object $O_t$, and the final action $a_m$ at step $l$ is to retrieve the target $O_t$, note that $O_{k_m} = O_{t}$.
Each action $a_i$ is represented as $(p_i,r_i,d_i)$, where $p_i \in \mathbb{R}^{3}$ denotes the grasp point on the manipulated object $O_{k_i}$, $r_i \in SO(3)$ represents the pose for grasping at $p_i$, and $d_i \in \mathbb{R}^{3}$ is the direction for retrieving $O_{k_{i}}$ after grasping at $p_i$.
% The manipulation sequence should be safe,
% meaning when manipulating each $O_{k_i}$ with $a_i$,
% other objects should not have displacements in their positions and poses.

% Further more, we define safely manipulation as \emph{no unplanned displacement}, which means when $A_{i_j}$ is implemented, object's displacement $S_t < \epsilon, t \in I\setminus\{i_j\}$, where $\epsilon$ is a small amount representing the motion threshold.
% Furthermore, we define safe manipulation as \emph{no unplanned displacement}, which means when $A_{i_j}$ is executed, the object's displacement $S_t < \epsilon, t \in I\setminus\{i_j\}$, where $\epsilon$ is a small amount representing the motion threshold.

%% file: tex/4_method.tex
\section{Method}

\subsection{Motivation and Overview}

As described in the \textbf{Introduction} section,
directly modelling the support relations between any two objects in clutters is difficult and inaccurate,
as object relations between two distant objects could be highly complicated and hard to predict because of the chained objects in between.

To tackle this problem,
we build the whole support relation graph of the cluttered objects by broadcasting the more accurate local dynamics between adjacent objects recursively (Section~\ref{sec:general} and~\ref{sec:clutter_solver}), with the assistance of \emph{Retrieval Direction Predictor} (Section ~\ref{sec:best_direction_proposer}) and \emph{Local Dynamics Predictor} (Section ~\ref{sec:local_dynamic_predictor}).
Guided by the support relation graph,
the robot can estimate the manipulation affordance (Section ~\ref{sec:affordance_estimator}) and execute the retrievals step by step.

\begin{figure*}[ht]
    \centering
    \includegraphics[width=18cm]{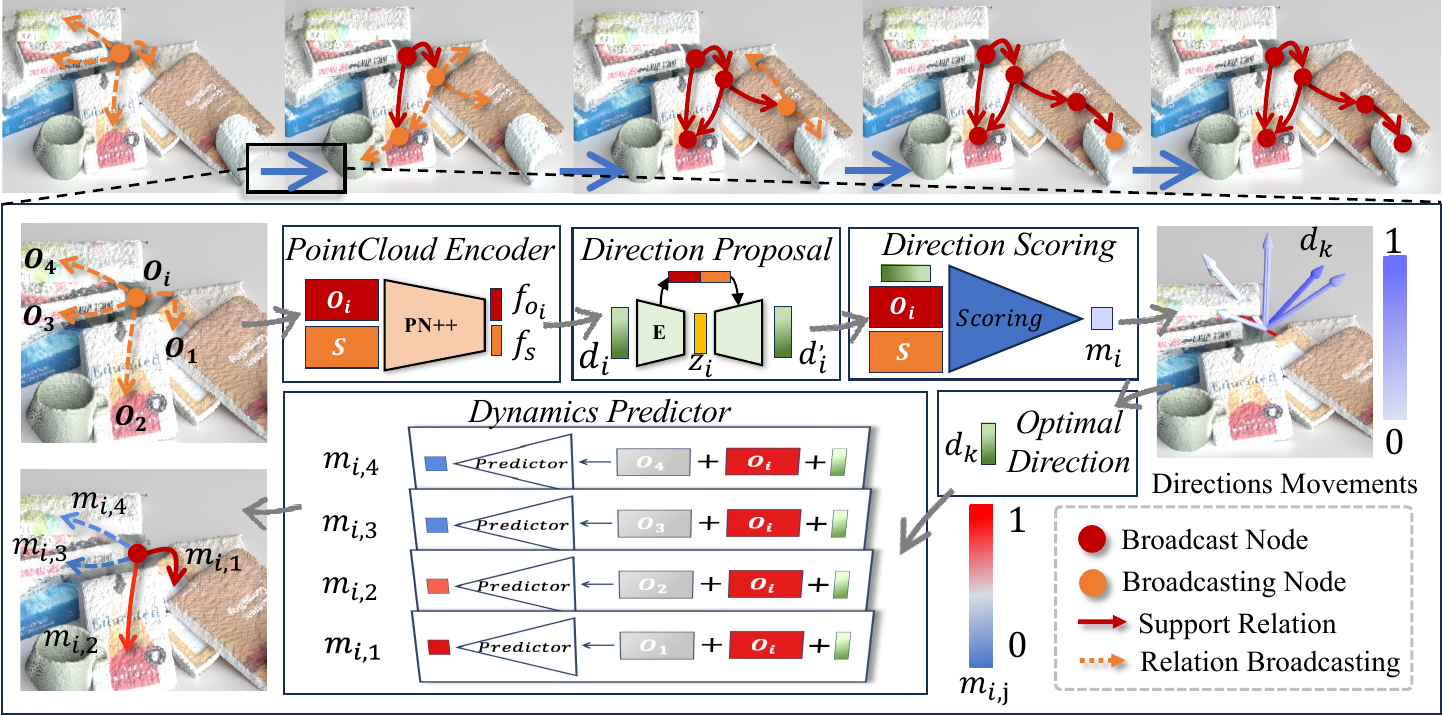}
    \caption{\textbf{Our Proposed Framework.} The first row shows the \textbf{Recursive Broadcasting} process of support relations via local dynamics. To infer the local dynamics starting from an object $O_i$, our framework first selects the optimal retrieval direction using the \emph{Direction Scoring Module} from the direction candidates proposed by the \emph{Direction Proposal Module}. With the optimal retrieval direction, the \emph{Dynamics Predictor} predicts the support relations between each object adjacent to $O_i$.}
    \label{fig:framework}
\end{figure*}

\subsection{General Idea for Support Graph Generation}
\label{sec:general}

To effectively represent complex scenes with multiple objects and their support relations, we adopt a directed acyclic graph (DAG) $\mathcal{G}$, where vertex $v_{i}$ represents object $O_i$ and edge $e_{ij}$ represents that object $O_i$ supports object $O_j$. Intuitively, to safely retrieve a target object $O_{t}$ without disturbing others, we need to estimate which objects are directly or indirectly supported by it. On the other hand, objects with no support relations to $O_t$ will minimally influence the retrieval process.
Therefore, our focus lies in the hierarchical support structure centered around the target object $O_t$, and we
construct a subgraph $\mathcal{G}_s$ that represents the support relations among cluttered objects, named as the \emph{Support Graph}.
This subgraph enables the derivation of a feasible retrieval sequence based on the spatial relationships outlined in the \emph{Support Graph}.

To set up the directed acyclic subgraph $\mathcal{G}_s$, there are three important steps:
\emph{First,} we build the nodes representing the objects in the clutter, ensuring that each object can be retrieved in a way that causes as less disturbance as possible. This necessitates optimizing the  retrieval direction to avoid collisions with other objects. Therefore, we propose the \emph{Retrieval Direction Predictor} (Section~\ref{sec:best_direction_proposer}), to generate and evaluate the optimal retrieval directions for each object. 
\emph{Subsequently}, we build the edges between different nodes, which indicate the presence of support relation between two objects. For this purpose, we introduce the \emph{Local Dynamics Predictor} (Section~\ref{sec:local_dynamic_predictor}), which predicts whether a given action on an object would cause another object to lose support and displace. \emph{Finally}, utilizing the two modules for building nodes and edges, we employ the \emph{Clutter Solver} (Section~\ref{sec:clutter_solver}) to recursively broadcast the inter-object relationships from the target object, thereby constructing the entire graph.

\subsection{Retrieval Direction Predictor} 
% \yan{How about this name? Retrieval Direction Predictor}
% \yan{How about using this name: Optimal Motion Predictor? since retrieving direction = motion direction} 
% I like this! \ruihai
% how about Optimal Retrieval Direction Predictor
\label{sec:best_direction_proposer}
When retrieving an object in clutter, different retrieval directions, denoted as $d \in \mathbb{R}^{3}$
could lead to different results, \emph{i.e.} different displacements of its adjacent objects. 
To minimize the displacements or collisions of adjacent objects, 
we introduce \emph{Retrieval Direction Predictor} to propose the optimal retrieval direction for the object that can avoid the movements of other objects to guarantee the safety and efficiency in manipulation for the less retrieval steps. 
% Specifically, the objectives of this module involves two scenarios:
% \begin{itemize}
%     \item for an object that can be retrieved without collisions, the module predicts an direction that avoids collision;
%     \item for an object that cannot be extracted without collisions, the module predicts the optimal retrieving direction, which will lead to minimal displacement of other objects.
% \end{itemize}

To achieve this goal, 
we propose two submodules in this predictor,
\emph{Direction Proposal Module} and \emph{Direction Scoring Module}.
The \emph{Direction Proposal Module} aims to propose direction candidates that will lead to minimal movements of other objects,
and the \emph{Direction Scoring Module} further scores the direction candidates, and select the best action direction.

% we initially use farthest-point-sample algorithm~\cite{qi2017pointnet++} in both 
In the \emph{Direction Proposal Module},
for each object point cloud $O_i$ and its scene point cloud $S$, we use PointNet++~\cite{qi2017pointnet++} to respectively extract their features $f_{O_i}$ and $f_{S}$.
We sample $q$ various retrieval directions $D=\{d_1,d_2,\dots d_q\}$ and obtaining the corresponding ground truth movement scores of other objects $\{m_1,m_2,\dots m_q\}$ (detail described in appendix).
For the retrieval directions with movement scores higher than a thresh $th_m$, we employ a conditional variational autoencoder (cVAE)~\cite{sohn2015learning} as the direction proposal model $F_{DP}$ to efficiently model the distribution of these good retrieval directions.
Specifically, the cVAE takes the feature concatenation of  $f_{O_i}$ and $f_{S}$ as the condition, encodes an retrieval direction $d_i$ to a latent $z_i$, and reconstructs $z_i$ into the direction $d_i^\prime$.
We employ the L1-loss between $d_i^\prime$ and $d_i$ as the reconstruction loss:
\begin{equation}
\label{eq1}
L_{recon}=|d_i-F_{DP}(f_{O_i}, f_{S}, d_i)|.
\end{equation}

A KL loss between $z_i$ and Gaussian distribution is applied to guarantee the sampling ability from an Gaussian noise $z$ to a direction in the promising directions distribution.

\begin{equation}
\label{eq2}
L_{KL}=\sum_{z_i}p(z_i)log(\frac{p(z_i)}{G(z_i)}),
\end{equation}

\noindent where $p$ means the distribution of $z_i$ and $G$ means the Gaussian distribution.

To further select the optimal retrieval direction, in the \emph{Direction Scoring Module},
we use Multi-Layer Perceptrons (MLPs) called $F_{DSM}$ to take the feature concatenation of $f_{O_i}$ and $f_{S}$ and $d_i$, and predict the corresponding movement score $\hat{m}_i$.
We employ the L1-loss between $\hat{m}_i$ and the ground truth movement ${m}_i$ as the loss:

\begin{equation}
\label{eq3}
L_{score}=|{m}_i-F_{DSM}(f_{O_i}, f_{S}, d_i)|.
\end{equation}

Such combination of Direction Proposal and Direction Scoring can progressively model the distribution of promising direction candidates and select the optimal retrieval direction leveraging the capability of two different-structured networks,
which is much better than directly proposing an action direction,
as promising actions lie in a distribution of multi-modal and dissimilar candidates but with similar scores, while a single direction prediction network could only generate single-modal predictions.

\subsection{Local Dynamics Predictor}
\label{sec:local_dynamic_predictor}
With the proposed optimal direction $d_k$ for an object $O_i$,
the next step is to estimate the resulting dynamics of other objects when $d_k$ is applied on $O_i$,  \emph{i.e.}, whether these objects will transition into a new state or remain static. 
However, due to the intricate relations between each objects within the clutter, 
directly predicting the dynamics state of each object in the clutter using a single network is both unfeasible and unnecessary. For example, the relations between two distant objects can be transferred by objects in between, which is quite a complicated process.

To address this challenge, we start from training \emph{Local Dynamics  Predictor}, aiming to estimate the adjacent object's dynamics state when a particular action is applied,
as thus relations are much easier for the model to make accurate inference.

Inspired by the particle-based dynamics predictor~\cite{chen2023predicting},
which uses particles to represent objects has demonstrated
great performance in modelling the dynamics of adjacent objects,
we use particle-based predictor to estimate the dynamics states of any object $O_j$ when retrieval is applied at direction $d_k$ on $O_i$, where $O_j$ is adjacent to $O_i$.
To be specific,
to extract the feature $g_{O_i}$
of each object $O_i$,
instead of directing encoding $O_i$ into a global feature,
we sample $r$ ($r=256$ in our paper) particles on $O_i$ and $O_j$ using farthest-point sampling, and merge the two groups of particles together with $d_k$ as the additional channel for $O_i$. Then we use Segmentation-version PointNet++ to extract per-point features of particles hierarchically and thus use the averaged features of sampled particles as the dynamics feature $g_{ij,k}$.
Then,
we use a Multi-layer Perceptron (MLP) called $F_{LDP}$ to take the dynamics feature $g_{ij,k}$ as input,
to predict the dynamics state of $O_j$.
To train this module, we employ 
Binary Cross Entropy (BCE) loss with ground truth labelled as 0 or 1 representing its dynamics state:
\begin{equation}
\label{eq4}
L_{dynamic}=BCE(Label,F_{LDP}(g_{ij,k})).
\end{equation}
\vspace{-2.5mm}

The \textbf{Experiments} section will show the empirical performance increase by using the particle-based dynamics model instead of directly extracting the object representations.
\begin{figure}[htb]
\centerline{
\includegraphics[width=1.0\linewidth]{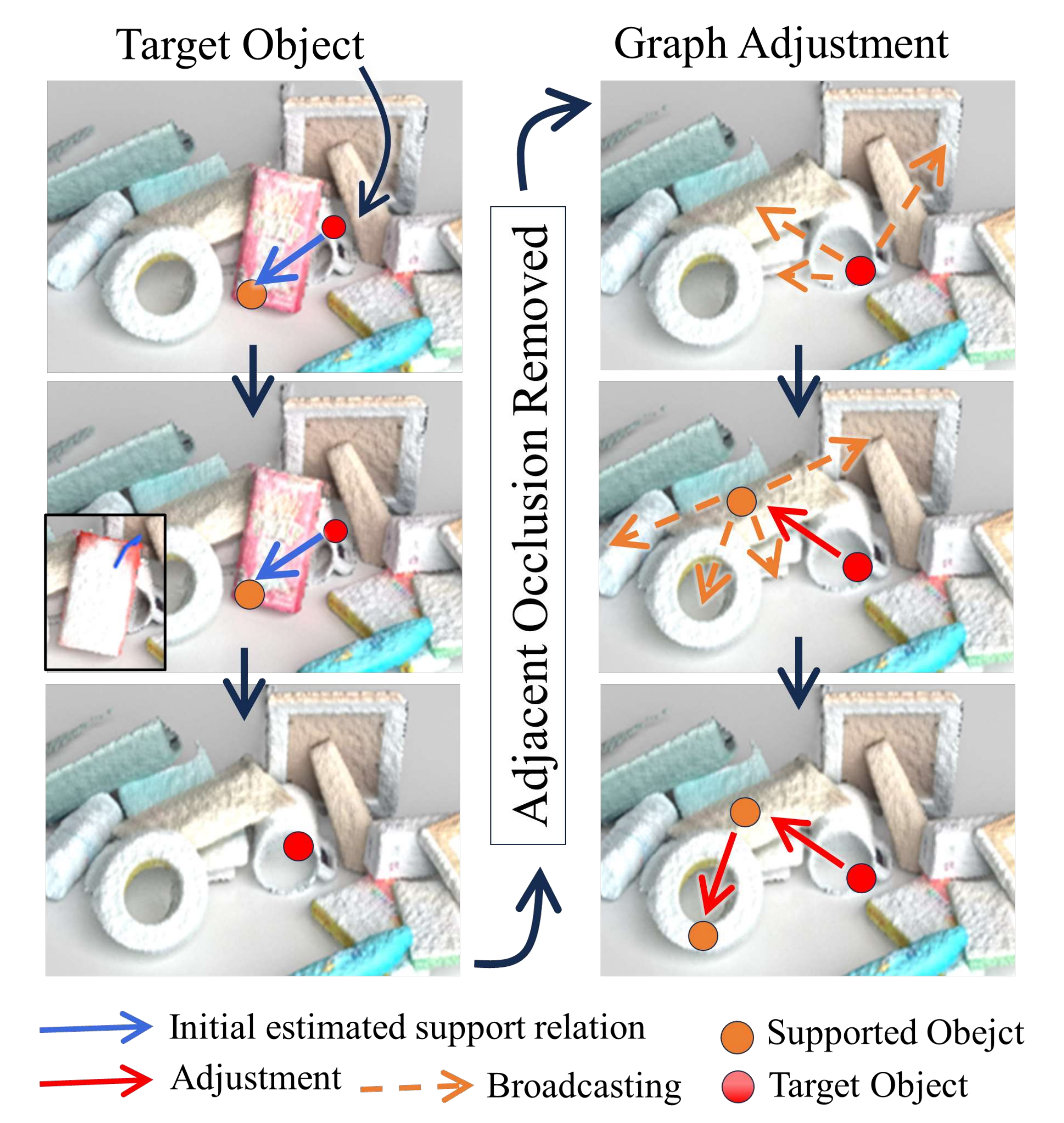}}
\caption{\textbf{Graph Adjustment} when the occlusion pink box is removed and the system re-broadcasts the supporting relations from the mug to be retrieved (target object).} \label{fig:ga}
\end{figure}

\subsection{Clutter Solver: Recursive Support Relation Broadcasting}
\label{sec:clutter_solver}
With the trained \emph{Retrieval Direction Predictor} and \emph{Local Dynamics Predictor},
for any object ${O_i}$ in \emph{Support Graph},
we can predict whether retrieving ${O_i}$ at the optimal retrieval direction will lead to the movement of its adjacent objects.
Specifically, we use \emph{Retrieval Direction Predictor} to propose the optimal retrieval direction $d_k$. 
Subsequently, for each $O_j \in N(O_i)$ (where $N(O_i)$ is the set of objects adjacent to $O_i$) we process them through \emph{Local Dynamics Predictor} respectively and obtain $m_{i,j}$ represent the movement of $O_j$ under the force of $O_i$ at its optimal retrieval direction $d_k$. 
We call this process the \emph{broadcast} of the dynamics of ${O_i}$. 
For $m_{i,j}$ larger than a threshold $th_m$, 
$O_j$ can be regarded as supported by $O_i$ and also the child node of $O_i$ in \emph{Support Graph}. 
It means to retrieve $O_i$ without collision, $O_j$ must be retrieved in advance. 
Besides, under the assumption that there is no mutually supportive relationship (such situation is quite rare and will require two robot arms to ensure the manipulation safety), 
we can ignore $O_i$'s parent nodes among $N(O_i)$ when broadcasting its dynamics.

Starting from target object $O_t$, and then its child nodes, we recursively use this mechanism to explore the \emph{Support Graph} and obtain the final graph after self-convergence. Note that not all support relations in the clutter is estimated because many of them are irrelevant to our goal of target retrieving. With \emph{clutter Solver} we can largely improve the computing efficiency in maximum extend than querying all support relations in the clutter in turn.

As the whole scene is partially observed, some objects which should be included in \emph{Support Graph} $\mathcal{G}$ may be ignored because the influence of occlusions.
Specifically,
they are actually supported by the target object, but currently occluded by other objects and thus temporarily could not exist in $\mathcal{G}$. 
However, when the occlusions are removed,
the supporting relation between the target and this object should be revealed.
To eliminate the impact of this case, we propose the \emph{Graph Adjustment} process to enhance the robustness for such a long-horizon task with occlusions.
Specifically, 
when an object $O_i$ is the next object to be retrieved, we conduct the broadcast again from $O_i$. 
When all the objects supported by $O_i$ all exist in $\mathcal{G}$, there is no need to further update $\mathcal{G}$.
However,
when there are novel objects supported by $O_i$ not existing in $\mathcal{G}$,
$\mathcal{G}$ should be recursively updated from the node of $O_i$.
Figure~\ref{fig:ga} gives a demonstration of this process.
The system originally estimates that only the pink box is supported by the target mug. when the pink box is removed,
before retrieving the mug,
the system first estimates the local dynamics of the mug and finds the previously hidden envelop is also support by the mug.
Then, the system further broadcasts the support relations from the envelop and recursively finds more support relations.

The benefit of \emph{Graph Adjustment} lies in two aspects. First, it can help to identify objects supported by $O_i$ which are wrongly inferred as non-supported objects before manipulation.
Because as the occluding objects is retrieved, more details of the hidden object are exposed and help to give a more accurate dynamics estimation.
Compared to inferring $\mathcal{G}$ from scratch (\emph{i.e.}, the target object), broadcasting the graph from $O_i$ only when a new object will exist in the $\mathcal{G}$ is much more cost efficient.

With the constructed \emph{Support Graph} $\mathcal{G}_{t_i}$ at each time step ${t_i}$,
only the objects in $\mathcal{G}_{t_i}$ should be considered to retrieve, while other objects don't have support relations with the target.
We can easily find one object $O_{t_i}$ in $\mathcal{G}_{t_i}$ that is directly or indirectly supported by the target object while not supporting other objects (\emph{i.e.}, the node $O_{t_i}$ with no outdegree in $\mathcal{G}_{t_i}$),
and retrieve it away by estimating the manipulation affordance.

\vspace{-2mm}

\begin{figure}[htb]
\centerline{
\includegraphics[width=1.0\linewidth]{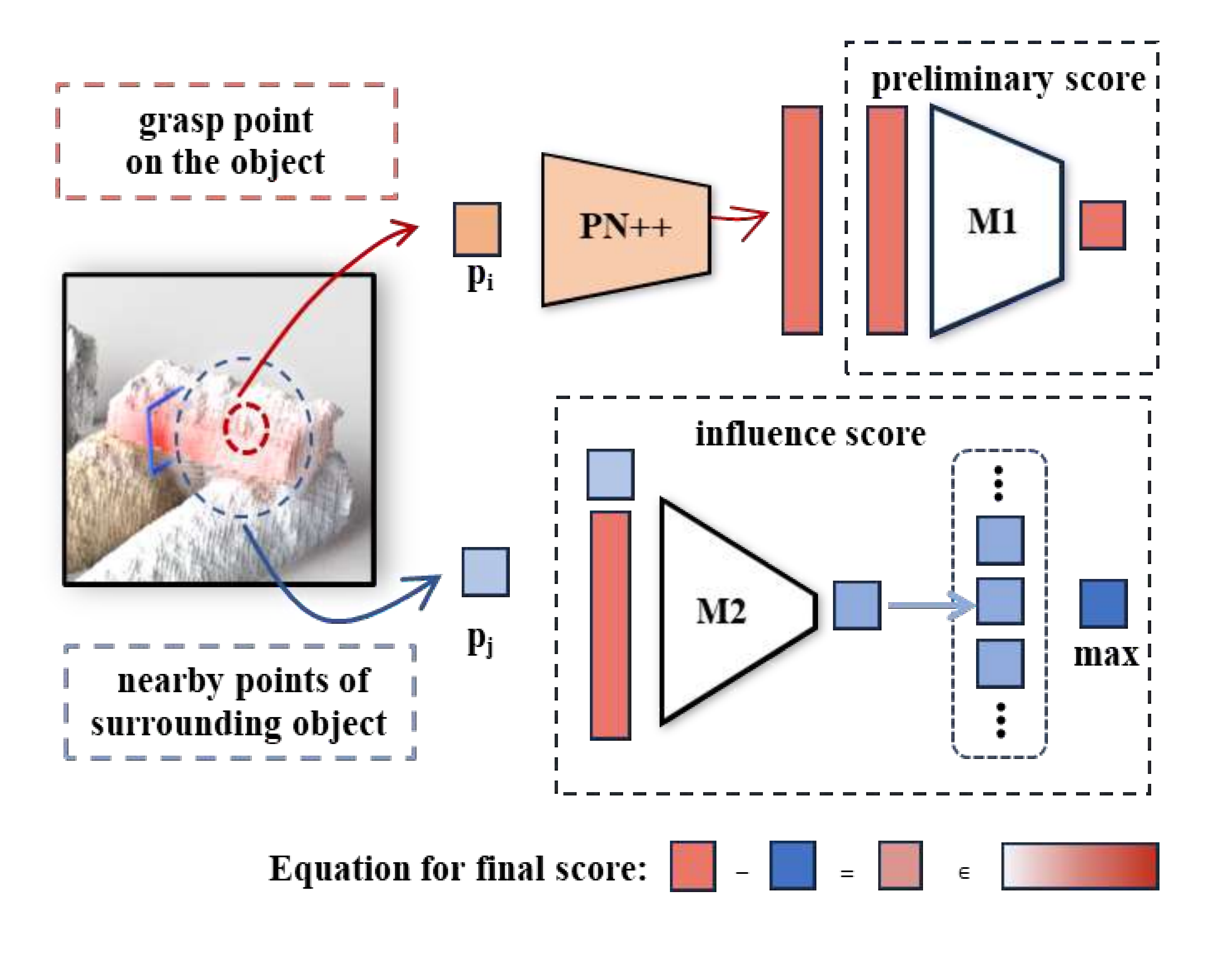}}
\caption{\textbf{Affordance Scoring Module}. To estimate the affordance score for a grasp point, we first calculate the preliminary affordance score solely based on the point itself, and then we evaluate the influence score by estimating the potential impact. The final affordance score is obtained by subtracting this influence score from the preliminary score.} \label{fig:eg}
\vspace{-6mm}
\end{figure}

\vspace{+0.5mm}

\subsection{Manipulation Affordance Predictor}
\label{sec:affordance_estimator}

% \cite{xxx} affordance papers

% the relationship with the previous sections
% the role of affordance module
% the design of module

After determining a feasible retrieval sequence through the generated \emph{Support Graph} and selecting the object $O_{k_i}$ to be manipulated at the current time step, our objective is to grasp this object while minimizing displacements or collisions with adjacent objects during manipulation.
Inspired by previous works~\cite{wu2021vat, xu2024naturalvlm, CVPR, zhao2022dualafford, ding2024preafford, wu2023learningforesightful, ling2024articulated} that demonstrated efficacy of visual affordance in offering generalizable actionable priors for diverse objects in 3D manipulation scenarios, we introduce the \emph{Manipulation Affordance Predictor} module. 
This module is designed to propose the optimal grasp point $p_i \in \mathbb{R}^{3}$ and grasp direction $r_i \in SO(3)$ to safely manipulate the object $O_{k_i}$ without disturbing others.

The \emph{Manipulation Affordance Predictor} consists of two submodules: the \emph{Affordance Scoring Module} and the \emph{Grasp Direction Predictor Module}. 
The \emph{Affordance Scoring Module} predicts the affordance score for difference grasp points, enabling the identification of high-quality grasp points and the selection of an optimal grasp point $p_i$ on the object.
Then, the \emph{Grasp Direction Predictor Module} generates the optimal grasp direction $r_i$ for the selected grasp point.

In the \emph{Affordance Scoring Module} (Figure~\ref{fig:eg}), the estimation of affordance scores for various points necessitates a comprehensive consideration of both the object's geometry and its surrounding environment. 
Initially, we calculate the preliminary affordance score for the grasp point $p_i$, excluding the influence of the surrounding clutter. Subsequently, we evaluate the potential impact of $p_i$ on the surrounding clutter, refining the preliminary affordance score. The final affordance score is then obtained by subtracting this refined impact from the preliminary affordance score.

To start with, we leverage PointNet++~\cite{qi2017pointnet++} to process the partial point cloud, extracting per-point features denoted as $f_{p_{i}}$. Then we use an MLP $M_1$ to decode the preliminary affordance score for each point based on its feature.
To estimate the potential influence a grasp point might exert on other objects, we identify all nearby points of $p_i$ from different objects, forming a set $A_{p_i}$, where for any point $p_j \in A_{p_i}$, $dist(p_i,p_j) < {\epsilon_x}$ and $p_j \notin O_{k_i}$. 
Then, we evaluate the potential impact of a grasp point $p_i$ on other object, by using an MLP $M_2$ to process both the grasp point feature $f_{p_i}$ and the position of the nearby point $p_j$ to decode the influence score. 
The final affordance score $s_p^i$ of $p_i$ is determined by subtracting the maximum influence score from the preliminary affordance score:
% and update the score of point to $p_j'$ condition on the environment 
\begin{equation}
s_p^i=M_1(f_{p_i}) - \underset{i \in A_{p_i}}{\max M_2( f_{p_i},p_j)}. 
\end{equation}

In the \emph{Grasp Pose Predictor Module}, similar to the method utilized in the \emph{Retrieval Direction Predictor} (Section~\ref{sec:best_direction_proposer}), we employ a conditional Variational Autoencoder (cVAE) to generate multiple candidates for grasp pose, and use Multi-Layer Perceptrons (MLPs) to predict the corresponding movement caused by these pose candidates. The final grasp pose is determined by selecting the poses which enable the robot manipulator to grasp and retrieve the object successfully without collision. 
The intricate architectural design and training strategy closely resemble those employed in the \emph{Retrieval Direction Predictor}, as outlined in Section~\ref{sec:best_direction_proposer}.

%% file: tex/5_experiment.tex
\section{Experiments}

\begin{figure*}[htpb]
    \centering
    \includegraphics[width=18cm, trim={0cm, 3cm, 0cm, 1cm}]{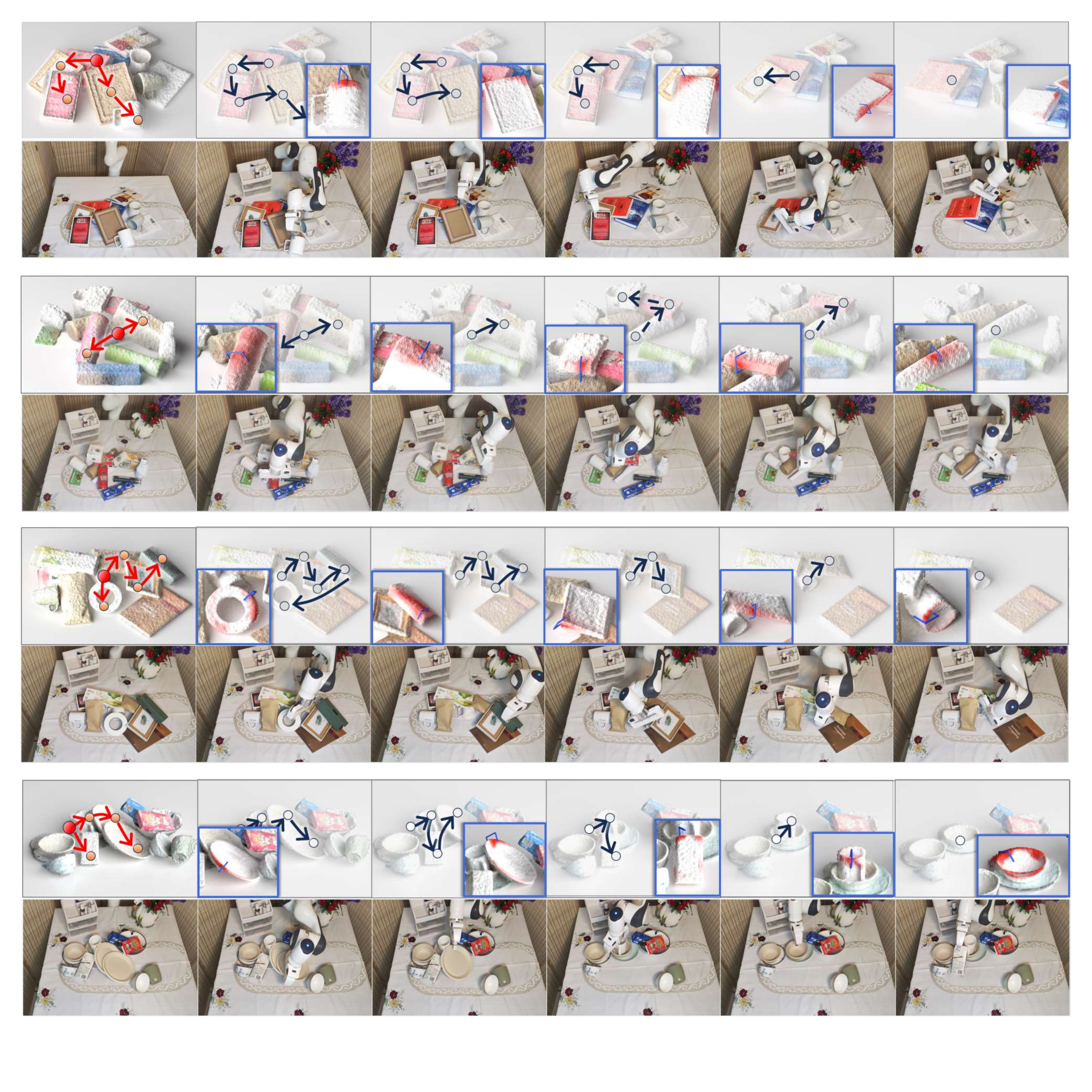}
    \caption{\textbf{Manipulation Sequence for Real-World Clutters with Captured Point Clouds.} We show the 4 cases respectively demonstrating the desk, food, sundries and kitchen scenarios. The second case contains occlusion removal and thus executes the \textbf{Graph Adjustment} process after moving away the white box in column 3.}
    \label{fig:result}
\vspace{-5mm}
\end{figure*}

\subsection{Setup}

For \textbf{simulation environment}, we equip OMNIVERSE ISAAC SIM~\cite{liang2018gpu} with 1 Franka Panda robot arm and 16 categories of thousands of different objects from ShapeNet~\cite{chang2015shapenet}, building up 4 different and realistic scenarios: kitchen, desk, food and sundries. The detailed data statistics is shown in section 1.1 in the appendix.
To train the manipulation policy,
we generate 5,000 different complicated clutters with different combinations of diverse objects for each scenario, where each clutter contains 15 objects on average.
This simulation environment with proposed dataset contains much more diverse objects and complicated clutters compared with previous evaluation environments~\cite{motoda2023multi,huang2023planning,xiong2021geometric}.

For \textbf{real-world setup},
we build the 4 scenarios (representative cases shown in the Figure~\ref{fig:result}), 
use Microsoft Azure Kinect (which has demonstrated high-precision with slight noises for robotic manipulation~\cite{ning2023where2explore}) to capture the point cloud of target scenes,
and Robot Operating System (ROS)~\cite{quigley2009ros} to control the Franka Panda robot arm for manipulation.
To capture the point cloud of each object,
we use Segment Anything (SAM)~\cite{kirillov2023segment} to segment each object, and project the corresponding depth image to the point cloud.

\subsection{Baseline, Ablations and Metrics}

To demonstrate the superiority of our framework in cluttered scenarios, we compare with 2 most recent strong \textbf{baselines}:
% , \textbf{RD-GNN}, which directly classifies relations between objects in the scene using Graph Neural Networks (GNN).
% \textbf{baselines}:
\begin{itemize}
    \item \textbf{RD-GNN}~\cite{huang2023planning} that directly classifies objects relations in the scene using Graph Neural Networks (GNN).
    \item \textbf{SafePicking}~\cite{Wada2022SafePickingLS} that uses object-level mapping and learning-based motion planning to achieve safe object retrieval. 
\end{itemize}

To demonstrate the effectiveness of different components of our framework, we compare with 4 \textbf{ablated versions}:

\begin{itemize}
    \item \textbf{Ours w/o DP} that removes the Retrieval Direction Predictor (DP) and always takes the upward direction for manipulation;
    \item \textbf{Ours w/o PR} that replaces the Particle-Based Representation (PR) with Object-level Representation for local dynamics prediction;
    \item \textbf{Ours w/o RB} that removes the Recursive Broadcasting (RB) process, and directly predicts the support relation between each 2 object;
    \item \textbf{Ours w/o GA} that removes the Graph Adjustment (GA) process and just uses the initial support map to conduct retrieval, without checking support relation after each manipulation step. 
\end{itemize}

For evaluation, we employ the following \textbf{metrics}:
\begin{itemize}
    \item \textbf{Retrieval Success Rate} that evaluates whether any displacement occurs during manipulation.
    \item \textbf{Accumulated Displacement Distance} demonstrates the mean of accumulated displacement distance for each scene in the test set during manipulation.
    \item \textbf{Retrieval Steps} that counts the average steps of retrieval under same successful retrieval cases.
    \item \textbf{Relation Prediction Success Rate} reflects whether the proposed dynamics states are correct.
    \item \textbf{Retrieval Direction Success Rate} that evaluates whether the proposed retrieval direction will lead to the least displacements of nearby objects.

\vspace{-2mm}
\end{itemize}

\vspace{-2mm}
\subsection{Results and Analysis}

\begin{figure*}[htpb]
    \centering
    \includegraphics[width=18cm]{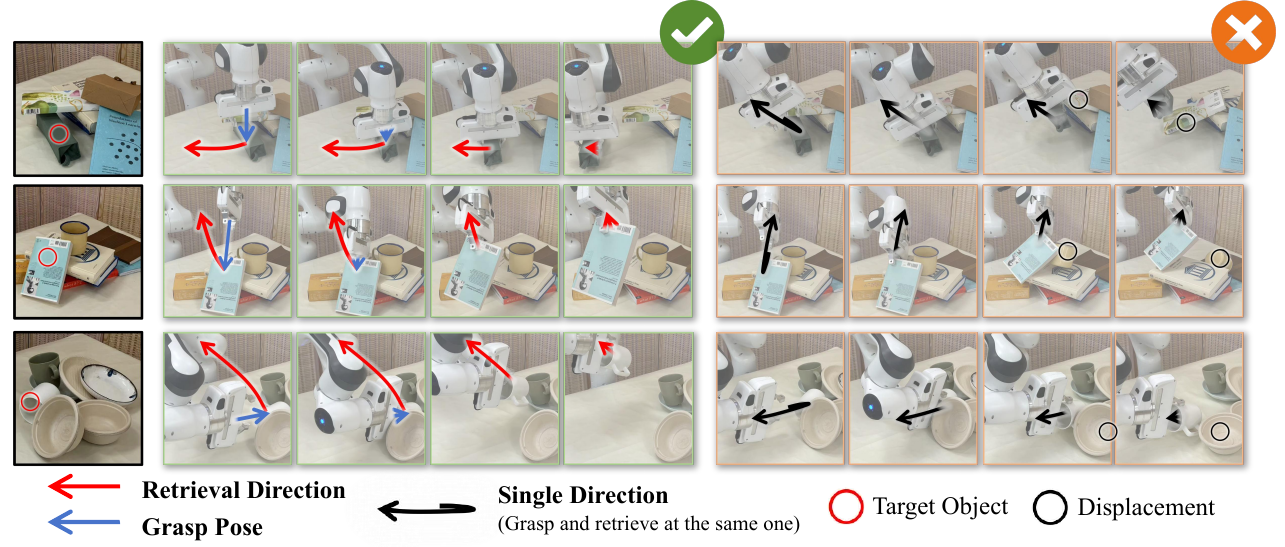}
    \caption{\textbf{Clarification between Grasp Pose and Retrieval Direction.} This figure reflects the manipulation flexibility brought by the separated \textbf{grasp pose} and \textbf{retrieval direction}. For example, the last row shows that the differentiation between \textbf{grasp pose} and \textbf{retrieval direction} enables the manipulator to reach optimality both in \textbf{grasp stability} and \textbf{collision avoidance with other objects} while the \textbf{single direction} easily leads to collision.}
    \label{fig:grasp_direction}
    \vspace{-2mm}
\end{figure*}

\begin{table}[!h]
    \centering
    \caption{\textbf{Clutter Object Retrieval Success Rate} which show overall success rate of our algorithm. It demonstrates that our design not only outperforms other algorithms but each part is crucial.}
    \resizebox{\linewidth}{!}{

    \begin{tabular}{lccccc}
        \toprule
         Method & Kitchen & Desk & Food & Sundries\\ \midrule
        Ours w/o DP& 0.72$\pm$0.038 & 0.79$\pm$0.041 & 0.73$\pm$0.032 &0.81$\pm$0.037\\ \midrule
        Ours w/o GA& 0.64$\pm$0.046 & 0.66$\pm$0.041 & 0.62$\pm$0.037 & 0.69$\pm$0.045\\ \midrule
        Ours w/o RB & 0.23$\pm$0.080 & 0.30$\pm$0.074 & 0.19$\pm$0.079 & 0.33$\pm$0.081\\ \midrule
        Ours w/o PR & 0.65$\pm$0.043 & 0.68$\pm$0.046 & 0.64$\pm$0.045 & 0.66$\pm$0.039\\ \midrule
        RD-GNN & 0.30$\pm$0.086 & 0.33$\pm$0.085 & 0.25$\pm$0.097 & 0.25$\pm$0.091\\ \midrule
        SafePicking & 0.37$\pm$0.085 & 0.42$\pm$0.078 & 0.34$\pm$0.081 & 0.35$\pm$0.077\\ \midrule
        \textbf{Ours} & \textbf{0.79$\pm$0.024} & \textbf{0.84$\pm$0.028} & \textbf{0.76$\pm$0.029} & \textbf{0.83$\pm$0.026}\\ \bottomrule
    \end{tabular}
    }
    \label{tab:success_sim}
\end{table} 

\begin{table}[!h]
    \centering
    \caption{\textbf{Retrieval Steps} under same successful cases. This criterion show that our algorithm achieves higher grasping efficiency compared to other baseline algorithms. Fewer number of grasping steps indicate higher efficiency.}
    \resizebox{\linewidth}{!}{
    \begin{tabular}{lccccc}
        \toprule
         method & Kitchen & Desk & Food & Sundries\\ \midrule
        Ours w/o DP& 5.21$\pm$0.66 & 4.75$\pm$0.71 & 5.13$\pm$0.62 & 5.62$\pm$0.67 \\ \midrule
        Ours w/o GA& 4.31$\pm$0.84 & 5.02$\pm$0.69 & 5.29$\pm$0.75 & 4.96$\pm$0.79 \\ \midrule
        Ours w/o RB& 7.95$\pm$1.13 & 7.72$\pm$1.09 & 7.91$\pm$1.21 & 7.75$\pm$1.08 \\ \midrule
        Ours w/o PR& 5.08$\pm$1.08 & 4.86$\pm$1.11 & 5.72$\pm$1.07 & 5.4$\pm$1.14 \\ \midrule
        RD-GNN & 7.41$\pm$1.38 & 7.83$\pm$1.27 & 7.56$\pm$1.32 & 6.67$\pm$1.43\\ \midrule
        SafePicking & 7.85$\pm$1.52 & 7.57$\pm$1.49 & 7.73$\pm$1.45 & 7.24$\pm$1.56\\ \midrule
        \textbf{Ours} & \textbf{4.11$\pm$0.52} & \textbf{4.61$\pm$0.43} &\textbf{4.95$\pm$0.54} & \textbf{4.51$\pm$0.49}\\ \bottomrule
    \end{tabular}
    }
    \label{tab:steps}

\vspace{-3.6mm}
\end{table}  

\begin{table}[!h]
    \centering
    \caption{\textbf{Accumulated Displacement Distance} shows the accumulated displacement distance for each scene (unit: centimeter). Lower total displacement Distance indicates the safer manipulation process.}
    \resizebox{\linewidth}{!}{

    \begin{tabular}{lccccc}
        \toprule
        method & Kitchen & Desk & Food & Sundries\\ \midrule
        Ours w/o DP& 9.44$\pm$0.86 & 8.81$\pm$0.94 & 9.63$\pm$0.0.90 & 9.58$\pm$0.91 \\ \midrule
        Ours w/o GA& 19.21$\pm$1.14 & 16.67$\pm$1.27 & 17.15$\pm$1.08 & 18.70$\pm$1.19 \\ \midrule
        Ours w/o RB& 27.33$\pm$2.74 & 24.96$\pm$2.45 & 28.21$\pm$3.07 & 25.78$\pm$2.51 \\ \midrule
        Ours w/o PR& 18.24$\pm$1.02 & 14.09$\pm$0.95 & 16.93$\pm$1.24 & 17.71$\pm$1.16 \\ \midrule
        RD-GNN & 23.49$\pm$1.71 & 23.57$\pm$1.96 & 26.78$\pm$1.85 & 28.14$\pm$1.43\\ \midrule
        SafePicking & 21.54$\pm$1.68 & 17.69$\pm$1.72 & 23.26$\pm$1.42 & 22.76$\pm$1.59\\ \midrule
        \textbf{Ours} & \textbf{6.57$\pm$0.77} & \textbf{5.89$\pm$0.64} &\textbf{6.21$\pm$0.79} & \textbf{6.04$\pm$0.73}\\ \bottomrule
    \end{tabular}
    }
    \label{tab:distance}

\vspace{-3.6mm}
\end{table}

Table~\ref{tab:success_sim}, ~\ref{tab:steps} and ~\ref{tab:distance}
show the large-scale evaluation results on the 3 main evaluation metrics in simulation.
For each scenario,
we set 357 different clutters for evaluation.
Our framework can infer the support graph of the scene, 
move away objects directly or indirectly supported by the target object step by step, and finally retrieve the target object safely.

The rapid performance increase from \textbf{RD-GNN}, \textbf{SafePicking}, \textbf{Ours w/o RB} (which all directly predict relations between each 2 objects) to \textbf{Our Whole Framework} in all the tables demonstrate that,
our main design,
\textbf{Recursive Broadcast},
is the fundamental mechanism for retrieving object in clutter,
as the support relation construction capability of the whole scene can be largely boosted by gradually broadcasting the more accurate local dynamics predictions. It is worth mentioning,
the second case shows the \textbf{Graph Adjustment} case. 
Specifically, while initially the framework cannot inference the hidden pink box is supported by the target object,
after retrieving the white box in the right,
the pink box is not occluded and our framework efficiently adjusts that it is supported by the target object, and should be moved away.
In this case,
the ablated version \textbf{Ours w/o GA} will not work, revealed in its performance decrease Table~\ref{tab:success_sim} and~\ref{tab:steps}.

\begin{figure}[htb]
\vspace{-3mm}
\centerline{
\includegraphics[width=0.8\linewidth]{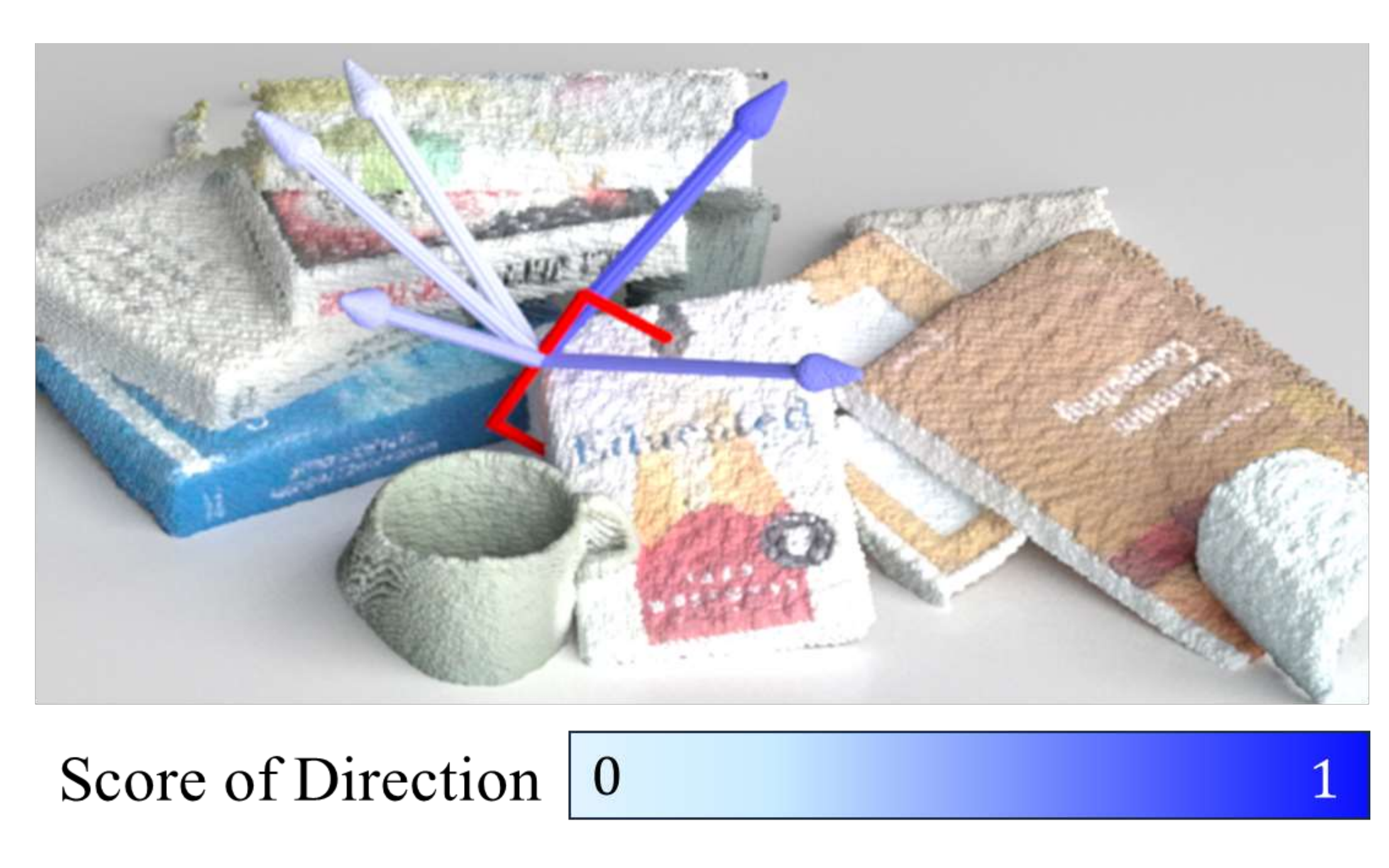}}
\vspace{-3mm}
\caption{\textbf{Scores of different directions} demonstrate that our framework proposes action directions that cause minimal disturbance to other objects. Higher scores mean better directions.} 
\vspace{-1mm}
\label{fig:direction}

\end{figure}

For \textbf{Retrieval Direction}, Figure~\ref{fig:grasp_direction} and Figure~\ref{fig:direction}  shows that our framework can effectively discriminate manipulation directions based on their potential collisions with other objects.
Besides,
the comparison between \textbf{Ours} and \textbf{Ours w/o DP} in Table~\ref{tab:direction} demonstrates that our method with the \emph{Retrieval Direction Predictor} will generate promising manipulation directions and thus leads to better performance.

\begin{table}[!h]
    \centering
    \caption{\textbf{Retrieval Direction Success Rate} demonstrates the significance of the \emph{Retrieval Direction Predictor}.}
    \resizebox{\linewidth}{!}{
    \begin{tabular}{lccccc}
        \toprule
         method & kitchen & desk & food & sundries\\ \midrule
        Ours w/o DP (no std) & 0.62 & 0.68 & 0.63 & 0.65 \\ \midrule
        \textbf{Ours} & \textbf{0.89$\pm$0.021} & \textbf{0.92$\pm$0.019} & \textbf{0.87$\pm$0.022} & \textbf{0.90$\pm$0.027}\\ \bottomrule
    \end{tabular}
    }
    \label{tab:direction}

\end{table}

For \textbf{Dynamics Prediction}, 
as shown in the comparison between \textbf{Ours} and \textbf{Ours w/o PR} in Table~\ref{tab:dynamics}, the particle-based representation, which is employed in our framework,
can highly improve the prediction accuracy compared with the object-level representation~\cite{chen2023predicting}.

\begin{table}[!h]
\vspace{-1.5mm}
    \centering
    \caption{\textbf{Relation Prediction Success Rate} shows the accurate dynamics prediction of \emph{Local Dynamics Predictor}.}
    \resizebox{\linewidth}{!}{
    \begin{tabular}{lccccc}
        \toprule
         method & Kitchen & Desk & Food & Sundries\\ \midrule
        Ours w/o PR & 0.73$\pm$0.052 & 0.78$\pm$0.050 & 0.74$\pm$0.053 & 0.77$\pm$0.058 \\ \midrule
        \textbf{Ours} & \textbf{0.89$\pm$0.029} & \textbf{0.93$\pm$0.028} & \textbf{0.90$\pm$0.031} & \textbf{0.91$\pm$0.028} \\ \bottomrule
    \end{tabular}
    }
    \label{tab:dynamics}
\end{table} 

Besides evaluating in simulator, we also conduct real world experiments. Table~\ref{tab:success_real} shows the success rate in the real world scenarios. 
In each scenario, we set 10 or 15 different clutters and execute the policy. 

\begin{table}[!h]
    \centering
    \caption{\textbf{Retrieval Success Rate} in the real world demonstrates the superior performance of our method compared to the baselines in real world clutter object retrieval tasks.}
    \resizebox{\linewidth}{!}{
    \begin{tabular}{lccccc}
        \toprule
         method & kitchen & desk & food & sundries\\ \midrule
        RD-GNN & 2/10 & 4/15 & 6/15 & 3/10\\ \midrule
        SafePicking& 4/10 &7/15 & 8/15 & 4/10 \\ \midrule
        \textbf{Ours} & \textbf{8/10} & \textbf{11/15} & \textbf{10/15} & \textbf{7/10}\\ \bottomrule
    
    \end{tabular}
    }
    \label{tab:success_real}
\end{table}  
\vspace{-1.5mm}

%% file: tex/6_conclusion.tex
\section{Conclusion}

In this paper,
we study the problem of cluttered objects manipulation.
As the clutter could be complicated and it is highly difficult to directly infer the support relations between any 2 objects,
we propose a framework broadcasting support relations recursively from local dynamics,
to effectively and efficiently predict the support graph of the whole clutter,
guiding the safe retrieval of the target object.
Extensive experiments showcase the superiority of our proposed framework.
% To retrieve a target object in the clutter without collisions with others,

\section{Acknowledge}
This paper is supported by National Natural Science Foundation of China - General Program (62376006), National Natural Science Foundation of China (No. 62136001),
The National Youth Talent Support Program (8200800081).